\documentclass{article}
\usepackage{iclr2027_conference}
\usepackage{lmodern}

\usepackage{amsmath,amsfonts}
\usepackage{hyperref}
\usepackage{url}
\usepackage{booktabs}
\usepackage[table]{xcolor}
\usepackage{makecell}
\usepackage{multirow}
\usepackage{graphicx}
\usepackage{cleveref}

\definecolor{First}{RGB}{102, 170, 255}
\definecolor{Second}{RGB}{166, 216, 245}
\definecolor{Third}{RGB}{225, 220, 245}

\newcommand{\first}[1]{\cellcolor{First}#1}
\newcommand{\second}[1]{\cellcolor{Second}#1}
\newcommand{\third}[1]{\cellcolor{Third}#1}

\newcommand{\illumTerm}{illumination term}
\newcommand{\IllumTerm}{Illumination term}
\newcommand{\emisTerm}{emission term}
\newcommand{\EmisTerm}{Emission term}
\newcommand{\globalGain}{global gain}
\newcommand{\spatialField}{spatial field}
\newcommand{\residual}{residual}

\title{PAM-ToD: Plug-and-Play Appearance Modeling for Cross-Time-of-Day 3D Gaussian Splatting}

\author{\normalfont
\begin{minipage}{0.97\textwidth}
\centering
 Kota Shimomura$^{1,2}$ Sungho Moon$^{3}$ \quad Tsubasa Hirakawa$^{1}$ \quad Takayoshi Yamashita$^{1}$\\[3pt]
Sunghoon Im$^{4}$ \quad Hironobu Fujiyoshi$^{1}$\\[7pt]
{\small $^{1}$Chubu University, Japan \qquad $^{2}$Elith Inc., Japan \\
$^{3}$DGIST, Daegu, Republic of Korea\
$^{4}$KAIST, Daejeon, Republic of Korea}
\end{minipage}}
\iclrfinalcopy
\begin{document}

\maketitle
\lhead{}
\renewcommand{\headrulewidth}{0pt}

\begin{abstract}
Adapting a pre-trained 3D Gaussian Splatting (3DGS) road scene to a new time of day requires learning appearance changes from a few anchor images while preserving consistent, real-time rendering.
We propose PAM-ToD, a lightweight plug-in that learns color corrections while keeping the pre-trained 3DGS parameters fixed.
PAM-ToD scales each Gaussian's existing color to model illumination changes and uses an additive term for additional brightness, such as when street lamps turn on at night.
Under a simplified image formation model, unchanged surface albedo can be eliminated from the relation between source and target appearances, allowing us to learn these corrections without separately estimating albedo and illumination.
The model corrects colors across the scene while allowing the corrections to vary by location and by Gaussian.
To guide learning from a few anchor images, it discourages abrupt spatial changes in these corrections.
We also introduce CARLA-ToD, a benchmark with matching geometry, camera poses, and moving-object trajectories across three times of day.
A few target-time anchor images are used to train each plug-in, while separate views are used for evaluation.
Across the static and dynamic settings, PAM-ToD achieves higher PSNR and lower LPIPS than the baselines, even when the anchor images come from a single synchronized capture across multiple cameras.

\end{abstract}

\section{Introduction}
Autonomous driving systems must operate reliably under changing lighting conditions, from dawn to night.
Evaluating them requires a simulator that can render the same road at different times of day.
3D Gaussian Splatting (3DGS)~\citep{kerbl2023gaussian} turns recorded drives into photorealistic scenes that can be rendered in real time~\citep{yan2024streetgaussians,zhou2024drivinggaussian,chen2025omnire}, supporting driving simulators and digital twins.
However, the reconstructed appearance remains tied to the lighting at capture time.
Reconstructing the road separately for each time of day repeats the data collection and training effort, increasing the cost of building a simulator.

A practical time-of-day simulator requires a relighting method that (i) reuses an already reconstructed 3DGS, (ii) relies only on sparse images of the new time of day, and (iii) preserves 3D consistency and real-time rendering.
Existing relighting approaches address these requirements with different trade-offs.
Inverse rendering approaches make 3DGS relightable by decomposing scene appearance into geometry, material, and illumination~\citep{gao2024relightable,liang2024gsir,herau2026}.
Recent methods further improve light transport modeling by incorporating global illumination~\citep{chen2025gigs} and spatially varying local lighting~\citep{du2025gsid}.
However, separating material and illumination remains ambiguous when the observations provide limited variation in lighting or viewpoint, especially in outdoor scenes.
Moreover, these methods typically construct a dedicated relightable representation rather than directly adapting an existing 3DGS to a new time of day.

Existing outdoor appearance models capture lighting variations through learned appearance embeddings or explicit illumination components~\citep{kulhanek2024wildgaussians,zhang2024gsw,kaleta2025lumigauss,bai2025gare,feng2025osdrgs,xiao2026}, but they are generally designed for image collections spanning multiple conditions.
Diffusion-based relighting offers flexible image and video editing~\citep{xing2025luminet,liang2025diffusionrenderer,he2025unirelight}; however, operating in image space without a shared 3D representation can compromise multi-view consistency, and iterative sampling is substantially slower than 3DGS rendering. These limitations motivate adapting a pre-trained 3DGS using only a few target-time images while preserving consistent, real-time rendering.

We address this need through \emph{Cross-Time Scene Transfer}, where a pre-trained 3DGS remains fixed and its appearance is adapted using a few posed target-time anchor images.
Our key idea is to learn appearance corrections directly, without separately recovering materials and illumination.
Under a simplified diffuse image-formation model, eliminating the time-invariant albedo yields a multiplicative illumination ratio and an additive correction that accounts for emission changes.
This relation provides the physical basis for adapting the source appearance directly, without explicitly recovering the underlying material and illumination.

Building on this formulation, we propose PAM-ToD, a lightweight appearance-adaptation plug-in for pre-trained 3DGS.
PAM-ToD models illumination variations across three levels of granularity, comprising a global scene-wide gain, a smooth spatial field, and a per-Gaussian residual.
All three levels are optimized jointly; capacity-proportional regularization encourages broad appearance changes to be explained by the global gain and the spatial field, while allowing local variation through the per-Gaussian residual.
Furthermore, an additive emission term captures newly active light sources, such as street lamps, which multiplicative scaling cannot generate in dark regions.
As a result, PAM-ToD achieves robust optimization even under sparse anchor image supervision.

Our main contributions are as follows:
\begin{itemize}
    \item \textbf{Cross-Time Scene Transfer.} We formulate appearance adaptation of a frozen, pre-trained road-scene 3DGS from a few posed target-time images.

    \item \textbf{PAM-ToD.}
    We introduce a plug-in module that combines hierarchical multiplicative corrections with additive brightness and regularization for sparse-anchor adaptation, enabling multiple time-of-day appearances without modifying the base 3DGS architecture.

\item \textbf{CARLA-ToD and Evaluation.}
    We construct a controlled benchmark with matched geometry, camera poses, and dynamic object trajectories across three times of day, demonstrating improved appearance transfer while preserving real-time rendering speed.

\end{itemize}

\section{Related Work}

\paragraph{Inverse Rendering.}
Inverse rendering approaches relighting by separating the contribution of a surface from that of the light falling on it.
Inverse rendering methods make this separation explicit by equipping Gaussian primitives with material and illumination attributes, where material parameters describe surface properties such as intrinsic color and reflectance.
Early methods such as Relightable 3D Gaussians~\citep{gao2024relightable}, GaussianShader~\citep{jiang2024gaussianshader}, and GS-IR~\citep{liang2024gsir} recover surface normals, material, and incident illumination under a physically based rendering model.
GI-GS~\citep{chen2025gigs} models indirect illumination, while GS-ID~\citep{du2025gsid} combines spatially varying local lighting with diffusion-based material priors.
As noted in the introduction, however, the separation remains ambiguous under the single illumination of a typical capture, and each method must reconstruct the scene in its own relightable representation.

\paragraph{Relighting for Outdoor Scenes.}
Outdoor captures of the same place differ widely with the time of day, the weather, and the shadows they produce.
GS-W~\citep{zhang2024gsw} models appearance variation through intrinsic and dynamic features, while WildGaussians~\citep{kulhanek2024wildgaussians} combines per-image and per-Gaussian embeddings to predict color transformations.
DAVIGS~\citep{lin2025davigs} further combines global appearance embeddings with 3D-consistent local features to model spatially varying color transformations.
These methods learn appearance variation from image collections rather than explicitly recovering illumination, making changes in lighting conditions only indirectly controllable through the learned appearance representation.
A different line of work extends inverse rendering to outdoor scenes. GaRe~\citep{bai2025gare}, OSDR-GS~\citep{feng2025osdrgs}, SU-RGS~\citep{zhang2025surgs}, and LumiGauss~\citep{kaleta2025lumigauss} explicitly model sunlight, skylight, and the shadows they cast.
For driving scenes, FEGR~\citep{wang2023fegr}, UrbanIR~\citep{lin2025urbanir}, and LightSim~\citep{pun2023lightsim} recover geometry, material, and sun and sky illumination from recorded drives, enabling relighting such as moving the sun or simulating night.
A multi-traversal method~\citep{xiao2026} instead separates material from illumination by recording the same route several times.
These approaches address appearance variation through learned appearance representations or explicit material and illumination models.
Our method instead derives corrections between source and target appearances and learns them as a plug-in on a frozen 3DGS, without separately estimating albedo and illumination.

\paragraph{Diffusion-based Relighting.}
A separate line of work changes illumination in images or videos with diffusion models.
For single images, IntrinsicDiffusion~\citep{luo2024intrinsicdiffusion} predicts intrinsic layers such as surface color and shading, IC-Light~\citep{zhang2025iclight} relights an image according to a text prompt or a background image, and DiffusionLight~\citep{phongthawee2024diffusionlight} estimates the lighting of a scene from one photograph.
Later work extends relighting to video, as in UniRelight~\citep{he2025unirelight}, UniLumos~\citep{liu2025unilumos}, and DiffusionRenderer~\citep{liang2025diffusionrenderer}, and to multiple viewpoints, as in LightSwitch~\citep{litman2025lightswitch}, while LumiNet~\citep{xing2025luminet} transfers the lighting of a reference image to an indoor scene.
Without an underlying 3D scene, nothing ensures that the same road looks the same from two viewpoints, and generating each frame takes many sampling steps, far more time than rendering a 3DGS.

\begin{figure}[t]
\centering
\includegraphics[width=\linewidth]{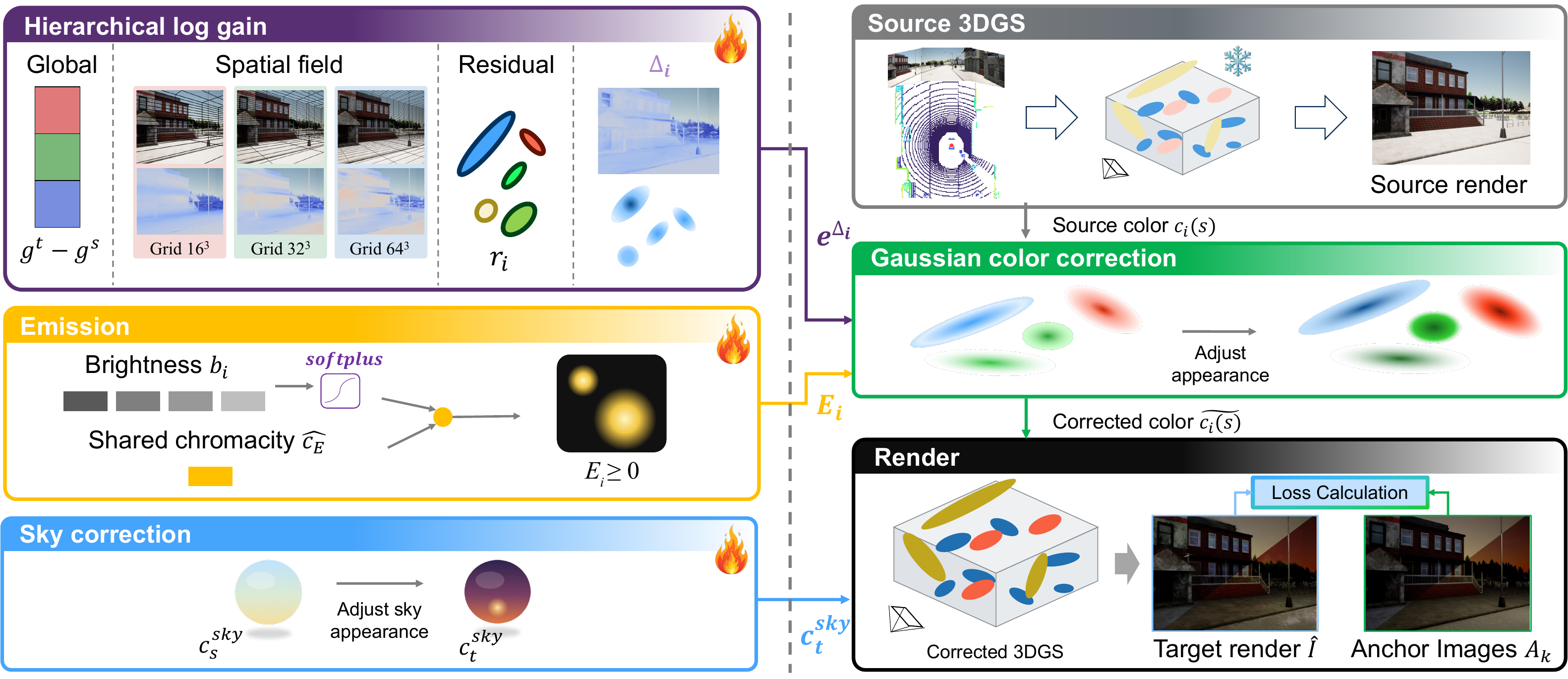}
\caption{\textbf{Overview of PAM-ToD}.Given a frozen 3DGS scene from source time $s$ and sparse anchor images at target time $t$, PAM-ToD adapts scene appearance in real time using a multiplicative \illumTerm{} and an additive \emisTerm{}.}
\label{fig:main_pipeline}
\end{figure}
\section{Method}
\label{sec:method}
As illustrated in Figure~\ref{fig:main_pipeline}, We propose \emph{PAM-ToD}, a render-time plug-in that adapts a pre-trained 3DGS scene to a different time of day while keeping all of its weights fixed.
Given a scene reconstructed at source time $s$ and posed anchor images captured at target time $t$, the plug-in adapts each Gaussian's color through a multiplicative \illumTerm{} and an additive \emisTerm{}.
We optimize the plug-in parameters using the anchor images, with implementation details provided in \cref{app:method}.

\subsection{Problem Formulation}
\label{sec:formulation}

\paragraph{Preliminaries.}
The source scene contains $M$ Gaussians $G=\{(\mu_i,q_i,\sigma_i,\alpha_i,c_i(s))\}_{i=1}^{M}$, where the attributes denote centers, rotations, scales, opacities, and source colors, respectively. It also includes a sky model $c_s^{\mathrm{sky}}(\omega)$, where $\omega$ is the viewing direction.
We freeze all Gaussian parameters and the sky model, and apply learned color corrections to their outputs at render time.
Each Gaussian carries a fixed class-probability vector $f_i$, averaged from SegFormer~\citep{xie2021segformer} predictions lifted from source images, and class label $\kappa(i)=\arg\max_c f_{i,c}$ (\cref{app:semantics}).
Our goal is to reconstruct the target-time appearance from arbitrary viewpoints using the source scene $G$ and a few anchor captures $\{A_k\}_{k=1}^{K}$, each containing synchronized images from the camera rig.

\paragraph{Appearance transport formulations.}
Motivated by diffuse reflection with additive emission, we adopt a simplified image formation model~\citep{kajiya1986rendering,1471710}, $c_i(t) = \rho_i \odot L_i(t) + E_i(t)$, where $\rho_i$ is a time-invariant albedo, $L_i (t)$ is an effective illumination factor, and $E_i (t)$ is an emitted radiance, all of which are three-channel and non-negative.
$\odot$ denotes channel-wise multiplication, and the ratios below are likewise taken channel-wise.
Assuming that the albedo is unchanged and the source illumination is non-zero in every channel, eliminating $\rho_i$ between the source and target equations yields
\begin{equation}
  c_i(t)
  =
  \underbrace{c_i(s) \odot \tfrac{L_i(t)}{L_i(s)}}_{\text{\illumTerm}}
  +
  \underbrace{E_i(t) - E_i(s) \odot \tfrac{L_i(t)}{L_i(s)}}_{\text{\emisTerm}}.
  \label{eq:transport}
\end{equation}
Eq.~\ref{eq:transport} expresses the target color as a rescaled source color plus an additive correction, without recovering albedo and illumination separately.
Note that the additive correction can be positive or negative even though both emissions are non-negative.

\subsection{Parameterization of the Transport Equation}
\label{sec:param}

Guided by Eq.~\ref{eq:transport}, PAM-ToD combines a multiplicative \illumTerm{} with log gain $\Delta_i\in\mathbb R^3$ and a nonnegative additive \emisTerm{} $E_i$, together with a separate sky correction.
Here, $c_i(s)$ is the view-dependent RGB of the source Gaussian obtained after SH evaluation, i.e., the color that would be rasterized without the plug-in (\cref{app:colorspace}).
At Gaussian center $x_i=\mu_i$ in world coordinates, the corrected color passed to the rasterizer is
\begin{equation}
  \tilde{c}_i = \mathrm{clamp}_{[0,1]}\!\Bigl(
      \underbrace{c_i(s) \odot e^{\Delta_i}}_{\text{\illumTerm}}
      + \underbrace{E_i}_{\text{\emisTerm}}
  \Bigr).
  \label{eq:model}
\end{equation}
Although the additive correction in Eq.~\ref{eq:transport} can be negative, we constrain $E_i$ to be non-negative by design: the gain $e^{\Delta_i}$ accounts for all darkening, including vanishing source emission, and $E_i$ accounts for additional brightness, including light sources that become active at the target time.
This rules out cancellation between an increased multiplicative contribution and a negative additive correction (\cref{app:sign}).%
The log gain $\Delta_i$ is parameterized hierarchically, with degrees of freedom that expand from a \globalGain{} to a \spatialField{} to a per-Gaussian \residual{}.
All components are optimized jointly; regularization strengthened in step with this capacity encourages broad appearance changes to be explained by the low-capacity \globalGain{} and \spatialField{}, while allowing local variation through the \residual{}.

\paragraph{\IllumTerm.}
We model the log illumination field at each time as $\hat{L}(x) = g + \sum_{l=1}^{3}\mathrm{trilerp}(G_l, x)$,
where $g \in \mathbb{R}^3$ is the \globalGain{}, $G_l \in \mathbb{R}^{3 \times n_l \times n_l \times n_l}$ are the multi-scale log-illumination grids that form the \spatialField{}, $\operatorname{trilerp}(G_l,x)$ denotes trilinear interpolation of $G_l$ at position $x$, and $l$ indexes three resolution levels with $n_l\in\{16,32,64\}$.
The log gain is the difference between the fields at times $t$ and $s$ plus a per-Gaussian \residual{} $r_i$:
\begin{equation}
  \Delta_i =
  \underbrace{(g^{t} - g^{s})}_{\text{\globalGain}}
  + \underbrace{\textstyle\sum_{l=1}^{3}\bigl[\mathrm{trilerp}(G^{t}_l, x_i) - \mathrm{trilerp}(G^{s}_l, x_i)\bigr]}_{\text{\spatialField}}
  + \underbrace{r_i}_{\text{\residual}}.
  \label{eq:delta-decomp}
\end{equation}
The \globalGain{} captures scene-wide brightness changes, while the \spatialField{} models smooth, spatially varying appearance changes such as sky gradients and large-scale shading variations across surfaces, with each grid regularized by a total-variation penalty that grows with its resolution.
The \residual{} $r_i \in \mathbb{R}^3$ is a directly optimized per-Gaussian parameter, initialized to zero, that absorbs local changes the \spatialField{} cannot capture; it is smoothed on a semantic $k$-NN graph whose edge weights $w_{ij}=\max(0,\cos(f_i,f_j))$ suppress smoothing across semantic boundaries (\cref{app:reg}).

\paragraph{\EmisTerm.}
The emission $E_i$ is a non-negative per-Gaussian vector that represents artificial light sources and local pools of light in dark regions.
To constrain emission color variation, we factor it into a learnable per-Gaussian brightness $b_i\in\mathbb{R}$ and a learnable chromaticity $\hat{c}_E\in\mathbb{R}_+^3$ of unit mean shared across the scene, $E_i = \mathrm{softplus}(b_i)\,\hat{c}_E$, initialized so that the emission is negligible and the plug-in reproduces the source render (\cref{app:init}).

\paragraph{Sky model.}
The frozen sky model is transported to the target time by a global affine transform, $c^{\mathrm{sky}}_t(\omega) = \mathrm{clamp}\bigl(e^{g_{\mathrm{sky}}} \odot c^{\mathrm{sky}}_s(\omega) + b_{\mathrm{sky}}\bigr)$ with the learnable $g_{\mathrm{sky}},b_{\mathrm{sky}}\in\mathbb{R}^3$ initialized to zero; the bias compensates for an absolute deficit of light in the source, as arises when transferring from night to day.

\subsection{Optimization}
\label{sec:supervision}

All plug-in parameters are optimized end-to-end by minimizing
$\mathcal{L} = \lambda_{\mathrm{anc}}\mathcal{L}_{\mathrm{anchor}} + \lambda_{\mathrm{atp}}\mathcal{L}_{\mathrm{atp}} + \lambda_{\mathrm{unseen}}\mathcal{L}_{\mathrm{unseen}} + \mathcal{L}_{\mathrm{chroma}} + \mathcal{L}_{\mathrm{reg}}$,
which combines a robust anchor reconstruction loss, an appearance transport prior that conditions the correction on the source color, a seen-to-unseen prior that propagates the adaptation from observed to unobserved regions, a chromaticity regularizer that suppresses hue noise, and structural regularizers matched to the representation hierarchy.
Every set-averaged term is defined as zero when its set is empty; full definitions are given in \cref{app:losses}.

\paragraph{Robust anchor reconstruction loss.}
The anchors $\{A_k\}$ are the only direct supervision, yet in practice they may be pseudo-images from a generative model, which may contain fine details that are inconsistent across views.
We therefore make the standard L1+SSIM loss robust to such anchors: the rendering $\hat{I}$ and the anchor $A$ are smoothed with a Gaussian filter $g_\sigma$ to reduce sensitivity to fine-scale differences while retaining broader appearance structure, and the set $\mathbb{T}_{\rho}$ of the fraction $\rho$ of pixels with the largest channel-averaged absolute errors after smoothing, is excluded from the L1 term:
\begin{equation}
  \mathcal{L}_{\mathrm{anchor}}
  = \operatorname*{mean}_{p \notin \mathbb{T}_{\rho}}
    \bigl\lvert (g_\sigma * \hat{I})_p - (g_\sigma * A)_p \bigr\rvert
  + \lambda_{\mathrm{ssim}}\bigl(1-\mathrm{SSIM}(g_\sigma * \hat{I}, g_\sigma * A)\bigr).
  \label{eq:robust-anchor}
\end{equation}
The sky correction receives gradients from this loss through sky pixels.

\paragraph{Appearance transport prior.}
Anchor pixels constrain only the Gaussians that project into them, whereas the time-of-day change of a surface is largely predictable from its semantic class and its source brightness: within a class, surfaces that are bright or dark at the source time map to consistently different target colors.
For this prior, we use view-independent source RGB values $c_i^{\mathrm{DC}}$ computed from the SH DC component.
For each class, the source colors and the anchor pixels are separately split into $Q$ luminance quantile bins, and each Gaussian receives as pseudo target $\tau_i$ the mean anchor RGB of the bin whose rank matches that of its source color.
The loss is an L1 pull toward these targets of the gain-corrected source color $\bar{c}_i$, which omits the emission so that the prior constrains the gain alone (\cref{app:losses}): $\mathcal{L}_{\mathrm{atp}} = \frac{1}{3|\mathbb{A}|}\sum_{i\in\mathbb{A}} \lVert\bar{c}_i-\tau_i\rVert_1$, over the set $\mathbb{A}$ of Gaussians with a valid $\tau_i$, whether visible from the anchors or not.

\paragraph{Seen-to-unseen transport prior.}
Gaussians outside every anchor view receive no gradient from $\mathcal{L}_{\mathrm{anchor}}$, so their log gain is left to whatever the smooth field extrapolates, which surfaces as inconsistent appearance in novel views; this prior transfers the fitted illumination change to them.
Let $\mathbb{V}$ be the set of Gaussians visible from at least one anchor view and $\mathbb{U}$ its complement.
Assuming that Gaussians of the same semantic class $\kappa$ and the same distance bin $d$ to the nearest light source (Gaussians labeled pole or traffic light serve as proxies) undergo a similar illumination change, we compute the per-channel median $m_{\kappa,d}$ of $\Delta_i$ over the visible members of each $(\kappa,d)$ cell, and during the second half of training pull the unseen set toward it,
$\mathcal{L}_{\mathrm{unseen}} = \frac{1}{|\mathbb{U}_{\mathrm{valid}}|}\sum_{i\in\mathbb{U}_{\mathrm{valid}}} \lVert\Delta_i-\operatorname{sg}(m_i)\rVert^2$ with $m_i \equiv m_{\kappa(i),d(i)}$,
where $\mathbb{U}_{\mathrm{valid}}\subseteq\mathbb{U}$ is the subset of unseen Gaussians for which a reference median $m_i$ is available (\cref{app:losses}), and $\operatorname{sg}$ denotes the stop-gradient operator; detaching the target makes the propagation one-directional and preserves the fit in the visible region.

\paragraph{Chromaticity regularization.}
A change of time of day alters brightness far more than hue, whereas the per-Gaussian \residual{} and the propagation to unseen Gaussians are free enough to introduce hue noise, visible as colored blotches on uniform surfaces such as roads.
We therefore penalize hue changes in the log gain while leaving luminance free.
With the chromaticity operator $\operatorname{ch}(v) = v - \frac{1}{3}\sum_{c} v_c$, the regularizer consists of three parts:
\begin{equation}
  \mathcal{L}_{\mathrm{chroma}}
  = \tfrac{\lambda_{\mathrm{rch}}}{3M} \sum_{i=1}^{M} \lVert\operatorname{ch}(r_i)\rVert^2
  + \tfrac{\lambda_{\mathrm{ch}}}{|\mathbb{U}_{\mathrm{valid}}|} \sum_{i \in \mathbb{U}_{\mathrm{valid}}} \lVert\operatorname{ch}(\Delta_i - \operatorname{sg}(m_i))\rVert^2
  + \tfrac{\lambda_{\mathrm{chu}}}{|\mathbb{C}|} \sum_{i \in \mathbb{C}} \lVert\operatorname{ch}(\Delta_i) - \operatorname{sg}(\tilde{h}_{\kappa(i)})\rVert^2.
  \label{eq:chroma}
\end{equation}
The first part restricts the \residual{} $r_i$ to luminance changes; the second suppresses spurious hue shifts when propagating to unseen Gaussians; and the third drives the chromaticity of uniformly textured classes $\mathbb{C}$ (road and sidewalk) toward $\tilde{h}_{\kappa}$, the detached per-class median of $\operatorname{ch}(\Delta_i)$ over $\mathbb{C}$ (\cref{app:losses}).

\paragraph{Structural regularization.}
Finally, because a few anchors constrain far fewer pixels than the plug-in has parameters, each level of the representation hierarchy receives one smoothness or sparsity regularizer matched to its capacity,
$\mathcal{L}_{\mathrm{reg}} = \lambda_{\mathrm{tv}}\mathcal{R}_{\mathrm{tv}} + \lambda_{\mathrm{knn}}\mathcal{R}_{\mathrm{knn}} + \lambda_{E}\mathcal{R}_{E}$,
where $\mathcal{R}_{\mathrm{tv}}$ is the resolution-weighted total-variation penalty on the \spatialField{} grids, $\mathcal{R}_{\mathrm{knn}}$ the semantic $k$-NN smoothness of the \residual{}, and $\mathcal{R}_{E}$ the $\ell_1$ sparsity penalty on the emission (\cref{app:reg}).
Optimizer settings and all hyperparameters are listed in \cref{app:hyper}.

\section{CARLA-ToD Dataset}
\label{sec:dataset}
Evaluating time-of-day adaptation pixel by pixel requires image pairs in which geometry, camera poses, and dynamic objects are held fixed and only the illumination changes.
Because real driving data cannot satisfy this condition, we construct the CARLA-ToD dataset with the CARLA simulator.
The dataset comprises 24 sequences (200 frames $\times$ 3 cameras each), covering four driving scenes at three times of day (Morning, Noon, Night) in two configurations (Static, Dynamic).
\vspace{-5pt}
\paragraph{Generation protocol.}
An ego vehicle is driven along a predefined route in synchronous mode with the same route and timing at every time of day, so the scene geometry and camera poses coincide exactly between corresponding frames (the difference in extrinsics is identically zero) and only illumination and emission change.
The weather is fixed to clear, and only the sun altitude is varied. Morning ($2.0^\circ$) reproduces the warm appearance of a low sun, Noon ($90.0^\circ$) a neutral illumination under a high sun, and Night ($-1.0^\circ$) an environment in which emission from artificial light sources, such as street lights and building windows, coexists with dark regions.
The Static configuration places no dynamic actors and is used to evaluate baselines that assume static scenes. The Dynamic configuration populates the scene with 40--60 vehicles and 40--55 pedestrians that follow identical trajectories at every time of day.
\vspace{-5pt}
\paragraph{Sensors and annotations.}
\label{sec:carla_mount}
The ego vehicle carries a rigid rig of three RGB cameras (Front, Front-left, Front-right), each recording $1920\times1300$ sRGB images with a horizontal field of view of $60^\circ$ and no lens distortion, and a roof-mounted 64-channel LiDAR that records about $1.1\times10^5$ points per frame. The mounting geometry is given in Appendix~\ref{app:dataset}.
We provide a binary sky mask for every RGB frame and, for the Dynamic configuration, binary masks of dynamic objects (all, vehicles, pedestrians) and 3D bounding boxes.

\section{Experiments}
To validate the proposed method, we evaluate on the newly constructed CARLA-ToD dataset and on the Waymo Open Dataset (WOD)~\citep{wod}.
Comparisons with inverse rendering and appearance-code baselines use the CARLA-ToD Static.
The proposed method uses StreetGS as its base model and is evaluated with $K=1$ and $K=8$ synchronized anchor captures.
Comparisons with diffusion-model baselines use the CARLA-ToD Dynamic setting, where each diffusion model is applied to the renderings of StreetGS.

\begin{table*}[t]
\small
\centering
\caption{
Quantitative relighting comparison on the CARLA-ToD Static Dataset.
The first, second, and third best performances are highlighted in
\colorbox{First}{First},
\colorbox{Second}{Second}, and
\colorbox{Third}{Third}, respectively.
}
\renewcommand{\arraystretch}{0.85}
\begin{tabular}{l|ccc|ccc|c}
\toprule
& \multicolumn{3}{c|}{\textbf{Reconstruction}}
& \multicolumn{3}{c|}{\textbf{Novel View Synthesis}}
& \textbf{FPS} $\uparrow$ \\
\textbf{Methods}
& SSIM $\uparrow$ & PSNR $\uparrow$ & LPIPS $\downarrow$
& SSIM $\uparrow$ & PSNR $\uparrow$ & LPIPS $\downarrow$
& \\
\midrule

GS-W
& \second{0.748}
& \third{16.76}
& \third{0.312}
& \third{0.729}
& \third{16.58}
& \third{0.324}
& 22.9 \\

GS-IR
& 0.663
& 16.42
& 0.372
& 0.671
& 16.57
& 0.361
& 61.8 \\

GI-GS
& 0.663
& 16.37
& 0.353
& 0.667
& 16.60
& 0.342
& 16.3 \\

LumiGauss
& 0.491
& 11.29
& 0.597
& 0.467
& 10.89
& 0.622
& 152.7 \\

\midrule

StreetGS
& 0.471
& 10.17
& 0.487
& 0.463
& 10.13
& 0.486
& 66.7 \\

+Ours ($K=1$)
& \third{0.740}
& \second{22.05}
& \second{0.298}
& \second{0.729}
& \second{21.89}
& \second{0.297}
& 53.4 \\

+Ours ($K=8$)
& \first{0.826}
& \first{25.66}
& \first{0.249}
& \first{0.814}
& \first{25.45}
& \first{0.247}
& 53.4 \\

\bottomrule
\end{tabular}
\label{tab:carla_comparison}
\end{table*}

\begin{figure}[t]
\centering
\includegraphics[width=\linewidth]{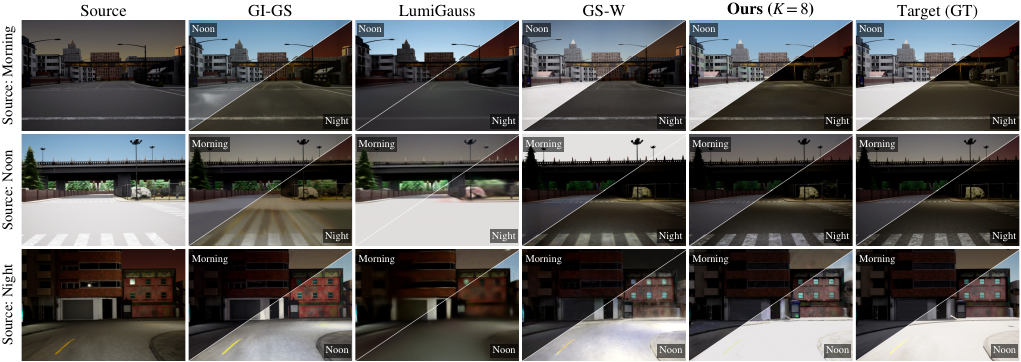}
\caption{Qualitative comparison on the CARLA-ToD Static Dataset.}
\label{fig:vis_ex1}
\end{figure}

\vspace{-8pt}
\paragraph{Implementation details.}
For each dataset, we trained all methods on a single NVIDIA A6000 GPU and compared their performance on 3D reconstruction and novel view synthesis (protocol details in \cref{app:dataset_splits}).
We further evaluated the contribution of each major component of our method by individually removing it and quantitatively measuring its impact.
Since GS-W requires images captured at different times of day in its training set, we trained it using a combination of eight frames from different times of day.
For GS-IR, GI-GS, and LumiGauss, we trained a separate model for each time of day and evaluated relighting by swapping the resulting environment maps across times of day.

\begin{figure}[t]
\centering
\includegraphics[width=\linewidth]{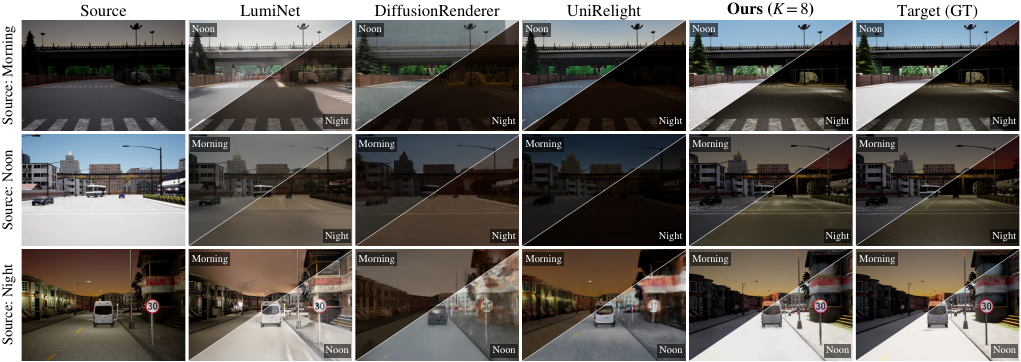}
\caption{Qualitative comparison on the CARLA-ToD Dynamic Dataset.}
\label{fig:vis_ex2}
\end{figure}

\vspace{-8pt}
\paragraph{Comparison on CARLA-ToD.}
Tables~\ref{tab:carla_comparison} and \ref{tab:carla_dynamic} report quantitative results on CARLA-ToD Static and Dynamic datasets, averaged over all six transfer directions (per-direction details in Appendix).
In Table~\ref{tab:carla_comparison}, our method achieves the best ToD adaptation quality while retaining real-time rendering.
On NVS ($K=8$), it achieves the highest accuracy, improving over baselines by up to 0.085 SSIM, 8.87 dB PSNR, and 0.077 LPIPS with negligible computational overhead, maintaining real-time rendering at 53.4 FPS.
Even with $K=1$, our physical inductive bias and hierarchical regularization suppress overfitting and significantly surpass baselines.
As shown in Figure~\ref{fig:vis_ex1}, inverse rendering methods struggle with ill-posed material decomposition, whereas appearance-embedding methods lack expressiveness for large illumination changes.
In Table~\ref{tab:carla_dynamic}, the proposed method ($K=1, 8$) likewise outperforms state-of-the-art diffusion-model methods.
Diffusion-model methods such as LumiNet and UniRelight operate in 2D image or video space and therefore cannot preserve 3D geometric consistency or photometric coherence across viewpoints.
Their multi-step denoising also makes inference extremely slow, at 0.07--0.08 FPS.
Figure~\ref{fig:vis_ex2} presents the qualitative comparisons.
LumiNet loses spatial consistency on continuous structures such as roads.
Meanwhile, Diffusion Renderer and UniRelight struggle to produce reference-faithful relighting and accurately capture fine details like road markings and moving vehicles.

\begin{table*}[t]
\small
\centering
\caption{
Quantitative relighting comparison on the CARLA-ToD Dynamic Dataset.
}
\renewcommand{\arraystretch}{0.85}
\begin{tabular}{l|ccc|ccc|c}
\toprule
& \multicolumn{3}{c|}{\textbf{Reconstruction}}
& \multicolumn{3}{c|}{\textbf{Novel View Synthesis}}
& \textbf{FPS} $\uparrow$ \\
\textbf{Methods}
& SSIM $\uparrow$ & PSNR $\uparrow$ & LPIPS $\downarrow$
& SSIM $\uparrow$ & PSNR $\uparrow$ & LPIPS $\downarrow$
& \\
\midrule

LumiNet
& 0.569 & 14.44 & \third{0.422}
& 0.560 & 14.45 & \third{0.421}
& 0.08 \\

DiffusionRenderer
& 0.541 & 14.49 & 0.496
& 0.529 & 14.49 & 0.566
& 0.08 \\

UniRelight
& \third{0.597} & \third{15.89} & 0.484
& \third{0.560} & \third{15.49} & 0.577
& 0.07 \\

\midrule

StreetGS
& 0.467 & 10.54 & 0.486
& 0.462 & 10.51 & 0.481
& 59.3 \\

+Ours ($K=1$)
& \second{0.715}
& \second{21.29}
& \second{0.318}
& \second{0.706} & \second{21.15} & \second{0.311}
& 49.4 \\

+Ours ($K=8$)
& \first{0.803}
& \first{24.79}
& \first{0.269}
& \first{0.791} & \first{24.45} & \first{0.263}
& 49.4 \\

\bottomrule
\end{tabular}
\label{tab:carla_dynamic}
\end{table*}

\vspace{-8pt}
\paragraph{Ablation on the number of anchors.}
Table~\ref{tab:k_scaling} varies the number of synchronized anchor captures $K$.
All metrics improve monotonically with $K$, with diminishing gains that saturate around $K=8$.
Even a single synchronized capture is within about 4 dB of $K=16$, indicating that the low-degree-of-freedom \globalGain{} and \spatialField{} explain most of the time-of-day change.
The remaining gap stems from regions unobserved by the anchors or saturated near zero in the dark source, where the \emisTerm{} is constrained only by anchor coverage.
\vspace{-8pt}
\paragraph{Ablation on the loss terms.}
Table~\ref{tab:loss_ablation} evaluates the contribution of individual loss terms under fixed structural regularization.
Removing the anchor reconstruction loss causes a noticeable drop in accuracy; nevertheless, performance remains above the base model as the appearance transport prior captures global shifts from per-class luminance statistics.
Omitting the appearance transport prior slightly degrades all metrics, validating its complementary effect.
The seen-to-unseen prior shows a minor effect here, given that eight synchronized anchor captures yield dense coverage and leave only a small unobserved set $\mathbb{U}$.
Finally, removing chromaticity regularization maintains PSNR but degrades SSIM and LPIPS, as it explicitly suppresses subtle hue blotches on flat surfaces that impact perceptual quality more than raw pixel errors.
\vspace{-8pt}
\paragraph{Ablation on the correction channels and hierarchy level.}

Table~\ref{tab:model_designs} shows that both correction channels are essential. Removing the multiplicative channel causes the largest drop at both anchor counts, whereas omitting the additive term yields smaller degradation.
The hierarchy levels also exhibit anchor-dependent behaviors.
At $K{=}8$, global gain captures most illumination changes, with spatial grids providing further gains and residuals offering marginal value.
At $K{=}1$, however, global gain alone is insufficient, making each hierarchy level necessary.
Although the residual-only variant approaches the full model in PSNR at $K{=}1$, its worse LPIPS and lower score at $K{=}8$ confirm that multi-scale grids are critical for cross-view consistency.
Overall, the full model consistently achieves the best performance, validating our multi-scale hierarchy under sparse anchor supervision, where global correction alone is insufficient.

\begin{table}[t]
\centering
\begin{minipage}[t]{0.5\textwidth}
  \centering
  \scriptsize
  \setlength{\tabcolsep}{4pt}
  \renewcommand{\arraystretch}{0.85}
  \caption{Ablation on the number of $K$.}
  \label{tab:k_scaling}
  \begin{tabular}{c|ccc}
  \toprule
  $K$ & SSIM $\uparrow$ & PSNR $\uparrow$ & LPIPS $\downarrow$ \\
  \midrule
  1  & 0.720 & 21.11 & 0.327 \\
  2  & 0.741 & 21.94 & 0.323 \\
  4  & 0.772 & 23.48 & 0.314 \\
  8  & 0.808 & 24.72 & 0.285 \\
  16 & \textbf{0.825} & \textbf{25.23} & \textbf{0.273} \\
  \bottomrule
  \end{tabular}
\end{minipage}%
\hfill
\begin{minipage}[t]{0.5\textwidth}
  \centering
  \scriptsize
  \setlength{\tabcolsep}{4pt}
  \renewcommand{\arraystretch}{0.85}
  \caption{Ablation on the loss terms.}
  \label{tab:loss_ablation}
  \begin{tabular}{l|ccc}
  \toprule
  Loss & SSIM $\uparrow$ & PSNR $\uparrow$ & LPIPS $\downarrow$ \\
  \midrule
  w/o $\mathcal{L}_{\mathrm{anchor}}$ & 0.622 & 15.05 & 0.429 \\
  w/o $\mathcal{L}_{\mathrm{atp}}$    & 0.803 & 24.54 & 0.289 \\
  w/o $\mathcal{L}_{\mathrm{unseen}}$ & 0.808 & 24.71 & 0.286 \\
  w/o $\mathcal{L}_{\mathrm{chroma}}$ & 0.804 & 24.75 & 0.291 \\
  Full & \textbf{0.808} & \textbf{24.72} & \textbf{0.285} \\
  \bottomrule
  \end{tabular}
\end{minipage}
\end{table}

\begin{table}[t]
\centering
\scriptsize
\setlength{\tabcolsep}{3pt}
\renewcommand{\arraystretch}{0.85}
\caption{Ablation on the correction channels and hierarchy levels with $K{=}1, 8$.}
\label{tab:model_designs}
\begin{tabular}{l|ccc|ccc}
\toprule
 & \multicolumn{3}{c|}{$K = 1$} & \multicolumn{3}{c}{$K = 8$} \\
Method & SSIM $\uparrow$ & PSNR $\uparrow$ & LPIPS $\downarrow$ & SSIM $\uparrow$ & PSNR $\uparrow$ & LPIPS $\downarrow$ \\
\midrule
Base & 0.471 & 10.17 & 0.487 &0.471 & 10.17 & 0.487 \\
w/o Illumination term & 0.565 & 12.83 & 0.431 & 0.659 & 16.73 & 0.375 \\
w/o Emission term & 0.717 & 20.86 & 0.329 & 0.783 & 22.69 & 0.311 \\
Global Only & 0.597 & 16.91 & 0.439 & 0.786 & 23.86 & 0.303 \\
Residual Only & 0.714 & 20.60 & 0.399 & 0.790 & 23.58 & 0.289 \\
Global + Sparse Grid & 0.689 & 20.01 & 0.358 & 0.804 & 24.70 & 0.289 \\
Full & \textbf{0.720} & \textbf{21.11} & \textbf{0.327} & \textbf{0.808} & \textbf{24.72} & \textbf{0.285} \\
\bottomrule
\end{tabular}
\end{table}

\begin{figure}[ht]
\centering
\includegraphics[width=\linewidth]{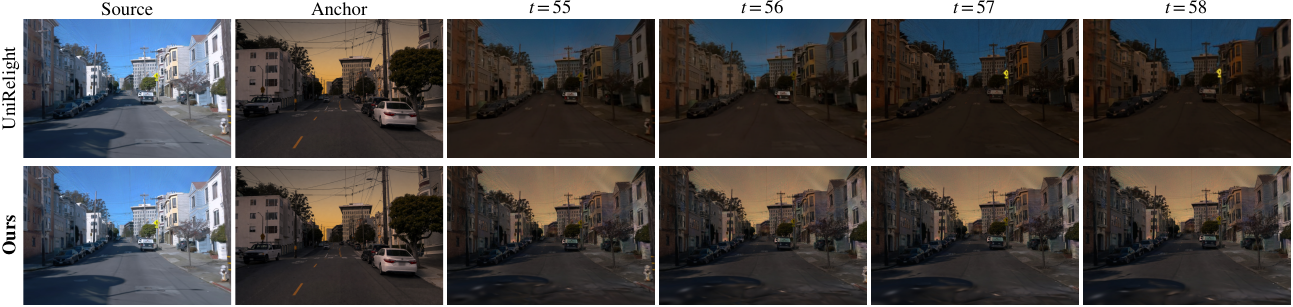}
\caption{Qualitative comparison on the Waymo Open Dataset.}
\label{fig:vis_ex3}
\end{figure}

\vspace{-10pt}
\paragraph{Real-Scene Adaptation on Waymo with Synthesized Anchors.}
We additionally evaluate whether PAM-ToD can adapt a 3DGS trained from WOD.
For this experiment, we use 4scenes as the source domain and synthesize sparse target-time anchor images using the OpenAI Image Edit API~\citep{openai}.
Figure~\ref{fig:vis_ex3} presents the qualitative comparison with UniRelight.
Across all reference-conditioned criteria, our method outperforms UniRelight, achieving lower errors in $\Delta E_{00}$ (2.69 vs.\ 8.10), $\Delta L^*$ (1.78 vs.\ 5.26)~\citep{sharma2005ciede2000}, and $\mathrm{EMD}_L$ (1.10 vs.\ 4.01)~\citep{rubner2000earth}. UniRelight suffers from degraded reference fidelity and chunk-wise brightness inconsistencies, whereas our approach faithfully retains target illumination. However, our method remains susceptible to floaters inherited from the source 3DGS, which diffusion-based models naturally eliminate as image enhancers.

\vspace{-5pt}

\section{Conclusion}
\vspace{-5pt}
We formalize Cross-Time Scene Transfer, adapting a frozen pre-trained road-scene 3DGS from sparse target-time images, and propose PAM-ToD as a lightweight plug-in.
Motivated by the cancellation of time-invariant albedo, PAM-ToD models appearance changes using a multiplicative correction and a non-negative additive emission term, avoiding explicit material-illumination decomposition.
Additionally, its hierarchical parameterization and near-identity initialization ensure stable adaptation even from a single synchronized anchor.
On CARLA-ToD, it substantially outperforms inverse rendering, appearance embedding, and diffusion baselines while maintaining real-time rendering.
\textbf{Limitations and Future Work.}
Our method requires posed anchor images at the target time of day.
Additionally, it struggles to capture specularities, sharp shadow boundaries, and physically consistent shadow rendering.
Future work includes extending our method to integrating physical shadow priors to handle severe illumination extremes.

\bibliography{references}
\bibliographystyle{iclr2027_conference}

\newpage
\appendix
\section{Method Details}
\label{app:method}

This appendix collects the implementation details omitted from \cref{sec:method}.

\subsection{Semantic Labels and Anchors}
\label{app:semantics}

We run SegFormer~\citep{xie2021segformer} with Cityscapes classes on the source images and lift the per-pixel class probabilities to the Gaussians by visibility-weighted averaging over all source views, where the weight of a view combines the Gaussian's opacity with an occlusion test against the rendered depth.
The averaged distribution is the semantic feature $f_i$ and its arg-max is the class label $\kappa(i)$; averaging also resolves conflicting observations across views.
Both quantities are computed once and kept fixed during plug-in optimization.
An anchor $A_k$ is one synchronized capture of the three cameras (front, front-left, front-right) of the same rig as the source scene, so $K$ anchors provide $3K$ posed images.

\subsection{Color Space and Point of Application}
\label{app:colorspace}

The correction of Eq.~\ref{eq:model} is applied to the view-dependent RGB of each Gaussian obtained after spherical-harmonics evaluation and before rasterization, in the display-encoded color space of the training images; no linearization is performed.
Eq.~\ref{eq:transport} therefore serves as a structural approximation in that space rather than as a radiometric identity, which is also why we do not interpret its two factors as physically recovered illumination and emission.
Two color representations of the source Gaussians are thus used: $c_i(s)$, the view-dependent RGB after spherical-harmonics evaluation, which is what rendering corrects, and $c_i^{\mathrm{DC}}$, the view-independent RGB given by the degree-0 SH coefficient alone, which the appearance transport prior of \cref{app:losses} uses because its pseudo targets are per-Gaussian statistics computed once without a viewing direction.

\subsection{Sign of the Additive Correction}
\label{app:sign}

The additive correction in Eq.~\ref{eq:transport} can be negative, e.g., when a source-time emitter is switched off at the target time, whereas $E_i$ in Eq.~\ref{eq:model} is non-negative.
Since $\rho_i$, $L_i$, and $E_i$ are never observed separately, the split in Eq.~\ref{eq:transport} is not unique: for any source color $c_i(s)>\mathbf{0}$, a negative additive part can be absorbed into the multiplicative term by
\begin{equation}
  e^{\Delta_i}\leftarrow\min\bigl(e^{\Delta_i},\,c_i(t)\oslash c_i(s)\bigr),
\end{equation}
leaving the non-negative remainder $\max\bigl(\mathbf{0},\,c_i(t)-c_i(s)\odot e^{\Delta_i}\bigr)$, where $\oslash$ denotes channel-wise division.
For an unconstrained per-Gaussian source--target RGB pair, the non-negativity of $E_i$ therefore does not reduce the set of representable target colors; the additive term is strictly necessary only where $c_i(s)\approx\mathbf{0}$, in which case the remainder is non-negative.
This argument concerns a single color pair and does not extend to the parameterization of Eq.~\ref{eq:model}, where the emission chromaticity is shared across Gaussians and the gain is shared across viewing directions.
Fixing the sign removes the ambiguity between a large gain and a cancelling negative emission, which we found to destabilize optimization with few anchors.
The final clamp in Eq.~\ref{eq:model} plays no role in this argument.

\subsection{Parameterization and Initialization}
\label{app:init}

\paragraph{Residual.}
$r_i\in\mathbb{R}^3$ is a directly optimized per-Gaussian parameter, initialized to zero; it is regularized, not aggregated, on the $k$-NN graph of \cref{app:reg}.

\paragraph{Emission.}
$b_i\in\mathbb{R}$ is a learnable per-Gaussian scalar that controls the emission magnitude through $\mathrm{softplus}$; it is initialized to $b_i=-6$, so that $\mathrm{softplus}(b_i)\approx2.5\times10^{-3}$ and the plug-in initially reproduces the source render.
The shared chromaticity $\hat{c}_E=3\,\mathrm{softplus}(\theta_E)/\lVert\mathrm{softplus}(\theta_E)\rVert_1$ is a positive vector of unit mean with learnable $\theta_E\in\mathbb{R}^3$, initialized to $\hat{c}_E=\mathbf{1}$.

\paragraph{Sky.}
$g_{\mathrm{sky}},b_{\mathrm{sky}}\in\mathbb{R}^3$ are the only trainable sky parameters; both are unconstrained and initialized to zero, so the transform starts as the identity, while the source sky model itself remains frozen.

\subsection{Loss Definitions}
\label{app:losses}

\paragraph{Robust anchor loss.}
Let $e_p=\frac{1}{3}\sum_{c}\lvert(g_\sigma*\hat{I})_{p,c}-(g_\sigma*A)_{p,c}\rvert$ be the channel-mean residual after blurring.
$\mathbb{T}_{\rho}$ contains the fraction $\rho$ of pixels with the largest $e_p$ within the anchor image rendered at the current iteration (\cref{app:hyper}; $\rho=10\%$, $\sigma=2$), without any additional masks.
Trimming applies only to the L1 term; the SSIM term is evaluated on all pixels of the blurred images.

\paragraph{Appearance transport prior.}
For each class $\kappa$, the view-independent source colors $c_i^{\mathrm{DC}}$ of its Gaussians and the anchor pixels labeled $\kappa$ by 2D segmentation of the anchor images are separately sorted by luminance ($0.299R+0.587G+0.114B$) and split into $Q{=}32$ equal-count quantile bins.
Each Gaussian receives as pseudo target $\tau_i$ the mean anchor RGB of the bin whose rank matches that of $c_i^{\mathrm{DC}}$; an empty anchor bin inherits the mean of the nearest lower non-empty bin (the class-wide anchor mean if no such bin exists), and classes with fewer than $Q$ Gaussians or fewer than 2{,}000 anchor pixels are skipped.
$\tau_i$ is computed once before optimization and is a rank-matching pseudo target rather than a pixel correspondence.
The gain-corrected source color of \cref{sec:supervision} is $\bar{c}_i=\mathrm{clamp}_{[0,1]}(c_i^{\mathrm{DC}}\odot e^{\Delta_i})$, which omits the emission term, so $E_i$ receives no gradient from this prior; for efficiency, the loss is evaluated on a random subset of $\mathbb{A}$ at each step.

\paragraph{Seen-to-unseen prior.}
The visibility weight of Gaussian $i$ is the sum over the anchor views of (opacity) $\times$ (in-frame indicator) $\times$ (non-occlusion indicator against the rendered depth); $\mathbb{V}$ is the set of Gaussians with a positive weight and $\mathbb{U}$ its complement.
Since true light-source positions are unknown, we use as proxies the Gaussians whose lifted segmentation has its arg-max at \emph{pole} or \emph{traffic light}; $d(i)$ is the Euclidean distance from $x_i$ to the nearest proxy center, quantized with the fixed, logarithmically spaced edges $\{2,4,8,16,32\}$\,m into six bins.
The bins depend only on the frozen source model and are computed once; in a scene without any proxy, the class-only median $m_{\kappa}$ is used instead.
The medians $m_{\kappa,d}$ are first computed over the visible set at the midpoint of optimization and recomputed periodically thereafter; $\mathcal{L}_{\mathrm{unseen}}$ and the second part of $\mathcal{L}_{\mathrm{chroma}}$ are active only from the midpoint on.
A cell $(\kappa,d)$ with too few visible members for a reliable median falls back to $m_{\kappa}$, and a class with too few visible members receives no prior (\cref{tab:hyper} lists the thresholds).
We denote by $\mathbb{U}_{\mathrm{valid}}\subseteq\mathbb{U}$ the unseen Gaussians whose class or cell has a valid reference median; the sum and the normalization of $\mathcal{L}_{\mathrm{unseen}}$ and of the second part of $\mathcal{L}_{\mathrm{chroma}}$ run over $\mathbb{U}_{\mathrm{valid}}$, so that unseen Gaussians without a reference contribute neither to the loss nor to its normalization.

\paragraph{Chromaticity regularization.}
In Eq.~\ref{eq:chroma}, $\mathbb{C}$ is the set of Gaussians labeled road or sidewalk (a class with fewer than 100 Gaussians is excluded), and $\tilde{h}_{\kappa}$ is the per-channel median of $\operatorname{ch}(\Delta_i)$ over the members of $\mathbb{C}$ with class $\kappa$, recomputed from the current values at every step and detached.
All three parts act on log-gain quantities; the first part is averaged over all $3M$ entries of the residual, as $\mathcal{R}_{E}$ is over the emission.
In dynamic scenes, the dynamic-object Gaussians share the \globalGain{} and \spatialField{} with the static scene but carry their own \residual{} and emission, to which the first part of $\mathcal{L}_{\mathrm{chroma}}$, $\mathcal{R}_{\mathrm{knn}}$, and $\mathcal{R}_{E}$ are applied in the same way; the appearance transport, seen-to-unseen, and uniform-class priors act on the static Gaussians only.

\subsection{Regularizers}
\label{app:reg}

The three terms of $\mathcal{L}_{\mathrm{reg}}$ are
\begin{align}
  \mathcal{R}_{\mathrm{tv}}
  &= \sum_{u\in\{s,t\}}\sum_{l=1}^{3}\frac{n_l}{n_1}\,
    \operatorname*{mean}_{\text{axes},\,\text{cells},\,\text{channels}}\bigl(\nabla G^{u}_l\bigr)^2,
  \label{eq:tv}\\
  \mathcal{R}_{\mathrm{knn}}
  &= \frac{\sum_{i}\sum_{j\in\mathcal{N}_k(i)} w_{ij}\,\lVert r_i - r_j\rVert_2^2}
         {\sum_{i}\sum_{j\in\mathcal{N}_k(i)} w_{ij}},
  \qquad w_{ij}=\max\bigl(0,\cos(f_i,f_j)\bigr),
  \label{eq:knn}\\
  \mathcal{R}_{E}
  &= \frac{1}{3M}\sum_{i=1}^{M}\lVert E_i\rVert_1.
  \label{eq:emis-l1}
\end{align}
In Eq.~\ref{eq:tv}, $\nabla G_l$ is the per-channel finite difference between neighboring cells along each axis, squared and averaged over the three axes, all cells, and the three channels; the penalty is applied to the source and target grids separately rather than to their difference, and the weight $n_l/n_1$ penalizes finer levels more strongly.
In Eq.~\ref{eq:knn}, $\mathcal{N}_k(i)$ are the $k{=}8$ Euclidean nearest neighbors of the Gaussian centers and $f_i$ is the fixed semantic feature of \cref{app:semantics}; edges with $w_{ij}=0$ contribute nothing and edges between Gaussians of different classes typically carry small but non-zero weights, so smoothing across semantic boundaries is suppressed rather than removed.

\subsection{Optimization and Hyperparameters}
\label{app:hyper}

All plug-in parameters, i.e., the global gain, the spatial-field grids, the per-Gaussian residual, the emission, and the sky correction, are optimized jointly for 8{,}000 iterations with a single Adam optimizer at a learning rate of $10^{-2}$; no level of the hierarchy is frozen, introduced later, or trained with a separate schedule.
Each iteration renders one anchor image, cycling through the $3K$ anchor images in a fixed order.
The seen-to-unseen prior and the second part of $\mathcal{L}_{\mathrm{chroma}}$ are the only terms with a schedule: they are enabled at iteration 4{,}000, when their reference medians are first computed from the visible Gaussians.
\Cref{tab:hyper} lists the loss weights and the remaining constants; the same values are used for all scenes and directions.

\begin{table}[h]
\centering
\small
\setlength{\tabcolsep}{4pt}
\caption{Hyperparameters of PAM-ToD, shared by all experiments.}
\label{tab:hyper}
\begin{tabular}{l p{0.55\linewidth}}
\toprule
Loss weights & $\lambda_{\mathrm{anc}}=20$, $\lambda_{\mathrm{ssim}}=0.2$, $\lambda_{\mathrm{atp}}=1$, $\lambda_{\mathrm{unseen}}=0.1$, $\lambda_{\mathrm{rch}}=1$, $\lambda_{\mathrm{ch}}=1$, $\lambda_{\mathrm{chu}}=2$ \\
Regularizers & $\lambda_{\mathrm{tv}}=10^{-2}$, $\lambda_{\mathrm{knn}}=10^{-2}$, $\lambda_{E}=10^{-4}$ \\
Spatial field & 3 levels ($n_l\in\{16,32,64\}$), TV weight $n_l/16$ \\
Residual graph & $k=8$ neighbors \\
Robust anchor & $\sigma=2$ pixels, $\rho=10\%$ \\
ATP prior & $Q=32$ bins, classes with $<2{,}000$ anchor px or $<Q$ Gaussians skipped, $65{,}536$ samples/iter \\
Seen-unseen & edges $\{2,4,8,16,32\}$\,m, min.\ 500 Gaussians/cell, update every 200 iters \\
Emission init & $b_i=-6$, $\hat{c}_E=\mathbf{1}$ \\
\bottomrule
\end{tabular}
\end{table}

\begin{table}[ht]
\centering
\scriptsize
\setlength{\tabcolsep}{3pt}
\renewcommand{\arraystretch}{1.0}
\caption{Sensor mounting relative to the front camera, in a camera basis with $x$ forward, $y$ left, and $z$ up. Location in m, rotation in deg.}
\label{tab:cam_mount}
\begin{tabular}{lcc}
\toprule
Sensor & Loc $(x,y,z)$ & Rot $(\mathrm{roll},\mathrm{pitch},\mathrm{yaw})$ \\
\midrule
Front camera (ID 0)       & $(0,0,0)$ & $(0,0,0)$ \\
Front-left camera (ID 1)  & $(-0.045,0.116,-0.001)$ & $(-0.05,-0.58,55.00)$ \\
Front-right camera (ID 2) & $(-0.049,-0.070,0.001)$ & $(-0.55,-0.44,-55.00)$ \\
LiDAR                     & $(-1.541,0.021,-0.300)$ & $(-0.70,0.42,-0.01)$ \\
\bottomrule
\end{tabular}
\end{table}

\section{CARLA-ToD Datasets Details}
\label{app:dataset}
This appendix complements the dataset description of the main paper with the generation protocol, the sensor configuration, the time-of-day settings, and the evaluation splits of CARLA-ToD.

\subsection{Generation Protocol}
\label{app:dataset_protocol}

\paragraph{Route and timing.}
All sequences are generated deterministically in synchronous mode with a fixed simulation tick, and the sensors are time-synchronized at 10\,Hz; each sequence lasts 20\,s (200 frames).

\paragraph{Static and Dynamic configurations.}
The Static configuration spawns no actor other than the ego vehicle, so each sequence contains only the static geometry of the town, and its dynamic-object masks are empty.
The Dynamic configuration additionally spawns 40--60 vehicles and 40--55 pedestrians that move through the scene; their per-frame poses and box sizes are exported and coincide exactly across the three times of day, so corresponding Dynamic frames again differ only in illumination.
The ego route is the same in both configurations, so the Static and the Dynamic sequences of a town share their camera poses as well.

\paragraph{Time-of-day settings.}
The weather is fixed to clear and only the sun altitude angle is changed: $2.0^\circ$ for Morning, $90.0^\circ$ for Noon, and $-1.0^\circ$ for Night; all other weather parameters and the camera settings are shared by the three conditions.
With the sun below the horizon, the simulator switches on its artificial light sources (street lights and illuminated building windows), so Night combines dark unlit regions with local emitters, whereas Morning and Noon are lit by the sun alone and differ in exposure and color temperature.

\subsection{Sensor Mounting and Imaging Parameters}
\label{app:dataset_sensors}

\paragraph{Sensor rig.}
A rigid rig of three RGB cameras (Front, Front-left, Front-right) and a roof-mounted LiDAR is attached to the ego vehicle.
The two side cameras are yawed by $\pm55^\circ$, so the three $60^\circ$ fields of view cover about $170^\circ$ horizontally with a small overlap between neighboring cameras.
~\Cref{tab:cam_mount} gives the mounting geometry relative to the front camera; the per-sequence extrinsics are constant and are exported for every frame as camera-to-world matrices in the same basis.

\paragraph{Image formation.}
Each camera records $1920\times1300$ 8-bit sRGB images at 10\,Hz with a $60^\circ$ horizontal field of view; the cameras are ideal pinholes without lens distortion.
The intrinsics follow from the field of view and the image size,
\begin{equation}
  f_x=f_y=\frac{W}{2\tan(\mathrm{FOV}/2)},\qquad c_x=\frac{W}{2},\qquad c_y=\frac{H}{2},
\end{equation}
which gives $f_x=f_y\approx1662.77$, $c_x=960$, and $c_y=650$ for $W{=}1920$, $H{=}1300$, and $\mathrm{FOV}{=}60^\circ$, shared by the three cameras.

\paragraph{LiDAR.}
The LiDAR has 64 channels and runs at 10\,Hz with a range of 100\,m and a vertical field of view of $[-25^\circ,15^\circ]$; each scan contains about $1.1$--$1.2\times10^5$ returns with a per-point intensity and is exported together with its pose.
The scans are used only to initialize the Gaussians and the scene bounds of the base model; neither the plug-in optimization nor the evaluation uses them.

\paragraph{Masks and annotations.}
A binary sky mask, aligned with the RGB frame, is provided for every image.
The Dynamic configuration additionally provides per-frame binary masks of the dynamic objects (all, vehicles, and pedestrians) for every camera and, for every actor, per-frame 3D bounding boxes given as an object-to-world pose and a box size; the base model uses the boxes to separate the dynamic-object Gaussians from the static scene.

\subsection{Evaluation Protocol}
\label{app:dataset_splits}

\paragraph{Source training views.}
Each sequence contains 200 synchronized captures (600 images).
Under the \emph{Reconstruction} protocol the source 3DGS is trained on all 200 captures of the source time.
Under the \emph{NVS} protocol every eighth capture is held out, and the source 3DGS is trained on the remaining captures.

\paragraph{Target anchors.}
The $K$ anchor captures are taken at the target time at frames spaced uniformly along the route, $\lfloor (k+\tfrac12)\,200/K \rfloor$ for $k=0,\dots,K-1$ (frame 100 for $K{=}1$; frames $12,37,\dots,187$ for $K{=}8$).
Under the NVS protocol an anchor that falls on a held-out frame is moved to the nearest training frame, so no held-out view is ever observed by the plug-in.

\paragraph{Evaluation views.}
NVS is scored on the held-out captures only, which are unseen by both the source 3DGS and the plug-in.

\section{Additional Results}
\subsection{Additional Results on Model size.}
\begin{table}[t]
\centering
\small
\setlength{\tabcolsep}{5pt}
\caption{Model size on CARLA-ToD Static. }
\label{tab:model_size}
\begin{tabular}{l|rrr}
\toprule
Method & Gaussians  & Params & Size \\
\midrule
GS-IR & 1.44\,M & 122.9\,M & 492\,MB \\
GI-GS & 1.58\,M & 125.9\,M & 504\,MB \\
GS-W &  1.91\,M & 173.7\,M & 695\,MB \\
\midrule
StreetGS & 1.48\,M & 106.2\,M & 425\,MB \\
+\,Ours & 1.48\,M & 113.9\,M \,(+7.7\,M) & 456\,MB \,(+31\,MB) \\
\bottomrule
\end{tabular}
\end{table}

As shown in Table~\ref{tab:model_size}, PAM-ToD adds only four floats per Gaussian for the residual and emission terms alongside scene-shared spatial grids totaling 7\,MB, resulting in 7.7\,M parameters and representing a mere 7\% increase over the frozen StreetGS base model.
In contrast, inverse rendering methods assign normals, albedo, roughness, and metallic parameters to every Gaussian while additionally maintaining lighting and shading volumes. Furthermore, GS-W attaches per-Gaussian appearance features $F \in \mathbb{R}^{16(K+1) \times H \times W}$.
While StreetGS requires a separate 3DGS model for each time of day, our method attaches a lightweight 31\,MB plug-in per target time to a single frozen base, reducing the total footprint for three time slots from 1.3\,GB to 0.49\,GB.
Although conventional baselines enable relighting within a unified framework, they suffer from trade-offs between reconstruction quality and rendering speed.
On the other hand, our approach preserves high reconstruction fidelity and real-time rendering speed while minimizing model capacity and reusing pre-trained 3DGS assets.

\subsection{Results Details on CARLA-ToD Static.}
Tables~\ref{tab:recon_comparison_appendix} and~\ref{tab:nvs_comparison_appendix} present detailed results for time-of-day adaptation across individual time slots in scene reconstruction and novel view synthesis, respectively.
Existing inverse rendering methods achieve time-of-day adaptation without significant accuracy drops when transferring between morning and night.
However, they suffer severe quality degradation under transitions such as noon to morning or night, as well as night or morning to noon.
This indicates that conventional inverse rendering fails to predict consistent material properties required for physically based rendering across different times of day for the same scene, demonstrating that simply swapping environment maps is insufficient for faithful time-of-day adaptation.
Additionally, LumiGauss frequently suffers from optimization instability, failing to converge during training.
While GS-W using appearance embeddings outperforms inverse rendering baselines, it still lacks sufficient expressive capacity.
In contrast, our method consistently yields superior average performance even with a single synchronized anchor capture.

\begin{table*}[t]
\scriptsize
\centering
\setlength{\tabcolsep}{3.5pt}
\renewcommand{\arraystretch}{1.05}

\caption{
Quantitative comparison of reconstruction on the CARLA-ToD Static Dataset.
Mo, No, and Ni denote Morning, Noon, and Night, respectively.
}
\label{tab:recon_comparison_appendix}

\begin{tabular}{c|c|cccccc}
\toprule
\textbf{ToD}
& \textbf{Metric}
& \textbf{GI-GS}
& \textbf{GS-IR}
& \textbf{LumiGauss}
& \textbf{GS-W}
& \textbf{Ours ($K{=}1$)}
& \textbf{Ours ($K{=}8$)}
\\
\midrule

\multirow{3}{*}{Mo $\rightarrow$ No}
& PSNR $\uparrow$
& 11.91
& 12.37
& 7.50
& \third{19.47}
& \second{20.93}
& \first{24.03}
\\
& SSIM $\uparrow$
& 0.697
& 0.689
& 0.428
& \second{0.815}
& \third{0.777}
& \first{0.827}
\\
& LPIPS $\downarrow$
& 0.362
& 0.382
& 0.609
& \third{0.265}
& \second{0.251}
& \first{0.237}
\\
\midrule

\multirow{3}{*}{No $\rightarrow$ Mo}
& PSNR $\uparrow$
& \third{19.81}
& 18.65
& 7.66
& 15.63
& \second{25.30}
& \first{27.00}
\\
& SSIM $\uparrow$
& 0.736
& 0.707
& 0.441
& \second{0.823}
& \third{0.798}
& \first{0.858}
\\
& LPIPS $\downarrow$
& 0.306
& 0.318
& 0.587
& \third{0.287}
& \second{0.278}
& \first{0.234}
\\
\midrule

\multirow{3}{*}{Mo $\rightarrow$ Ni}
& PSNR $\uparrow$
& \third{19.97}
& 20.24
& 19.23
& 19.67
& \second{22.39}
& \first{26.93}
\\
& SSIM $\uparrow$
& 0.680
& 0.686
& 0.624
& \second{0.739}
& \third{0.737}
& \first{0.835}
\\
& LPIPS $\downarrow$
& 0.312
& 0.336
& 0.529
& \third{0.309}
& \second{0.289}
& \first{0.233}
\\
\midrule

\multirow{3}{*}{Ni $\rightarrow$ Mo}
& PSNR $\uparrow$
& \third{20.86}
& 20.46
& 19.13
& 19.13
& \second{23.77}
& \first{27.34}
\\
& SSIM $\uparrow$
& 0.681
& 0.699
& 0.627
& \third{0.731}
& \second{0.741}
& \first{0.832}
\\
& LPIPS $\downarrow$
& 0.317
& 0.337
& 0.550
& \second{0.292}
& \third{0.315}
& \first{0.263}
\\
\midrule

\multirow{3}{*}{No $\rightarrow$ Ni}
& PSNR $\uparrow$
& 15.93
& \third{16.14}
& 7.20
& 14.15
& \second{22.01}
& \first{26.24}
\\
& SSIM $\uparrow$
& 0.581
& 0.572
& 0.408
& \second{0.706}
& \third{0.704}
& \first{0.821}
\\
& LPIPS $\downarrow$
& 0.387
& 0.408
& 0.637
& \third{0.345}
& \second{0.324}
& \first{0.250}
\\
\midrule

\multirow{3}{*}{Ni $\rightarrow$ No}
& PSNR $\uparrow$
& 9.74
& 10.63
& 7.04
& \third{12.50}
& \second{17.93}
& \first{22.44}
\\
& SSIM $\uparrow$
& 0.602
& 0.628
& 0.415
& \third{0.675}
& \second{0.689}
& \first{0.786}
\\
& LPIPS $\downarrow$
& 0.433
& 0.452
& 0.667
& \third{0.374}
& \second{0.336}
& \first{0.281}
\\

\bottomrule
\end{tabular}
\end{table*}

\begin{table*}[t]
\scriptsize
\centering
\setlength{\tabcolsep}{3.5pt}
\renewcommand{\arraystretch}{1.05}

\caption{
Quantitative comparison of novel view synthesis (NVS) on the CARLA-ToD Static Dataset.
Mo, No, and Ni denote Morning, Noon, and Night, respectively.
}
\label{tab:nvs_comparison_appendix}

\begin{tabular}{c|c|cccccc}
\toprule
\textbf{ToD}
& \textbf{Metric}
& \textbf{GI-GS}
& \textbf{GS-IR}
& \textbf{LumiGauss}
& \textbf{GS-W}
& \textbf{Ours ($K{=}1$)}
& \textbf{Ours ($K{=}8$)} \\
\midrule

\multirow{3}{*}{Mo $\rightarrow$ No}
& PSNR $\uparrow$
& 12.74 & 12.98 & 7.51 & \third{16.41} & \second{20.70} & \first{23.56} \\
& SSIM $\uparrow$
& 0.719 & 0.715 & 0.400 & \second{0.780} & \third{0.757} & \first{0.806} \\
& LPIPS $\downarrow$
& 0.339 & 0.352 & 0.634 & \third{0.310} & \second{0.252} & \first{0.235} \\
\midrule

\multirow{3}{*}{No $\rightarrow$ Mo}
& PSNR $\uparrow$
& \third{20.42} & 19.11 & 7.99 & 15.65 & \second{25.04} & \first{27.00} \\
& SSIM $\uparrow$
& 0.744 & 0.731 & 0.437 & \third{0.780} & \second{0.788} & \first{0.848} \\
& LPIPS $\downarrow$
& \third{0.289} & 0.301 & 0.601 & 0.310 & \second{0.273} & \first{0.223} \\
\midrule

\multirow{3}{*}{Mo $\rightarrow$ Ni}
& PSNR $\uparrow$
& 19.72 & 20.19 & 19.03 & \third{20.97} & \second{22.30} & \first{26.60} \\
& SSIM $\uparrow$
& 0.672 & 0.684 & 0.600 & \second{0.724} & \third{0.724} & \first{0.820} \\
& LPIPS $\downarrow$
& 0.315 & 0.330 & 0.551 & \third{0.304} & \second{0.290} & \first{0.237} \\
\midrule

\multirow{3}{*}{Ni $\rightarrow$ Mo}
& PSNR $\uparrow$
& \third{20.93} & 20.56 & 19.28 & 19.02 & \second{23.53} & \first{27.09} \\
& SSIM $\uparrow$
& 0.680 & 0.698 & 0.597 & \third{0.723} & \second{0.729} & \first{0.820} \\
& LPIPS $\downarrow$
& \third{0.305} & 0.332 & 0.574 & \second{0.292} & 0.312 & \first{0.260} \\
\midrule

\multirow{3}{*}{No $\rightarrow$ Ni}
& PSNR $\uparrow$
& 15.95 & \third{16.17} & 7.21 & 15.46 & \second{21.94} & \first{26.10} \\
& SSIM $\uparrow$
& 0.581 & 0.578 & 0.410 & \second{0.711} & \third{0.697} & \first{0.812} \\
& LPIPS $\downarrow$
& 0.382 & 0.397 & 0.647 & \third{0.334} & \second{0.322} & \first{0.249} \\
\midrule

\multirow{3}{*}{Ni $\rightarrow$ No}
& PSNR $\uparrow$
& 9.84 & 10.40 & 6.88 & \third{11.99} & \second{17.85} & \first{22.33} \\
& SSIM $\uparrow$
& 0.604 & 0.618 & 0.399 & \third{0.654} & \second{0.677} & \first{0.774} \\
& LPIPS $\downarrow$
& 0.424 & 0.452 & 0.690 & \third{0.394} & \second{0.333} & \first{0.278} \\

\bottomrule
\end{tabular}
\end{table*}

\subsection{Qualitative Evaluation of Ablation Studies.}
\begin{figure}[ht]
\centering
\includegraphics[width=\linewidth]{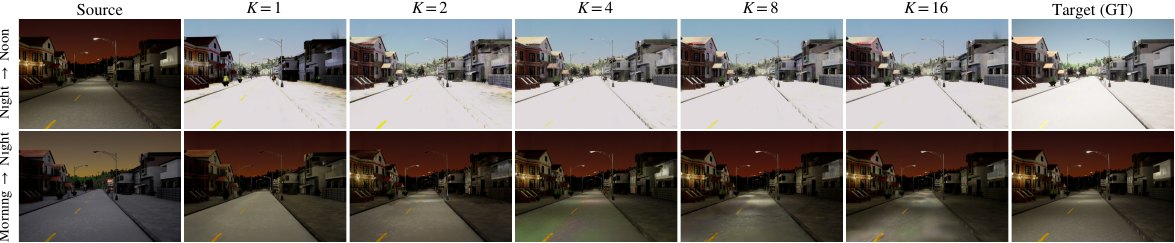}
\caption{Qualitative comparison on of the number of anchors.}
\label{fig:appenndix_1}
\end{figure}

\paragraph{Ablation on the number of anchors.}
~\Cref{fig:appenndix_1} shows the results of the comparison of the number of anchors.
All metrics improve monotonically with the number of anchor captures, with gains that diminish beyond $K{=}8$: at that density the anchors already see most of the scene, and further captures mostly re-observe Gaussians that are already constrained.
A single synchronized capture recovers most of the change, because the dominant part of a time-of-day transfer is a global change of exposure and color temperature, which the low-capacity \globalGain{} and \spatialField{} extrapolate to the whole scene from a single capture.
The benefit of additional anchors is strongly direction-dependent.
The dark regions improve more than the lit ones as anchors are added: the former are explained by the gain, whose field becomes better constrained, while the latter depend on the emission of the observed Gaussians.

\begin{figure}[ht]
\centering
\includegraphics[width=\linewidth]{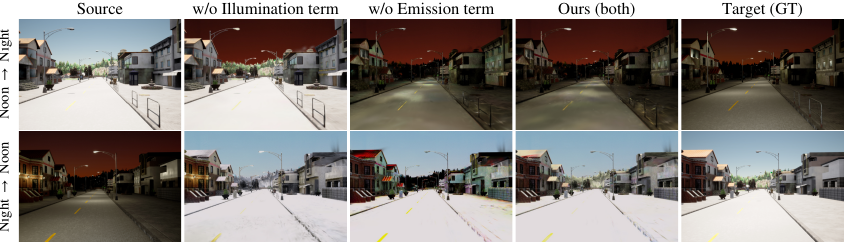}
\caption{Qualitative comparison on our key illumination and emission term.}
\label{fig:appenndix_2}
\end{figure}

\paragraph{Ablation on our key illumination and emission term.}
Figure~\ref{fig:appenndix_2} presents qualitative results of the ablation study on our key illumination and emission terms.
Removing these two components leads to fundamentally different failure modes.
Without the illumination term, appearance transport relies entirely on a non-negative offset, restricting the model to brightening the scene.
Consequently, transitions to darker times remain largely identical to the unadapted source, making it difficult to adjust local colors for brighter time slots because an offset merely shifts the color spectrum rather than scaling the underlying source texture.
Conversely, removing the emission term still allows the gain to capture exposure, color temperature shifts, and large-scale shading variations, which functions reasonably well when the target is darker than the source.
However, near-black source colors require large gains, which amplify their residual color noise into severe chromatic artifacts on road surfaces and walls; the additive term provides another way to increase brightness without such amplification.
In the displayed examples (Figure~\ref{fig:appenndix_2}), the decomposition into bright and dark regions is consistent with this functional separation: the emission term concentrates on active light sources and changes little in the remaining dark regions, so on its own it does not model fine local appearance shifts.
Nevertheless, when only a single synchronized capture is available, the contribution of the emission term remains limited because per-Gaussian degrees of freedom cannot be fully constrained from a single capture, allowing the low-capacity illumination term to absorb most of the appearance shift.
Overall, these two terms are non-redundant and mutually complementary, as an offset cannot darken a scene and a multiplicative gain cannot lift exactly zero channels, demonstrating that both components are indispensable for robust adaptation.

\subsection{Additional ToD-Dataset and Real-World Datasets Results.}
\begin{figure}[ht]
\centering
\includegraphics[width=\linewidth]{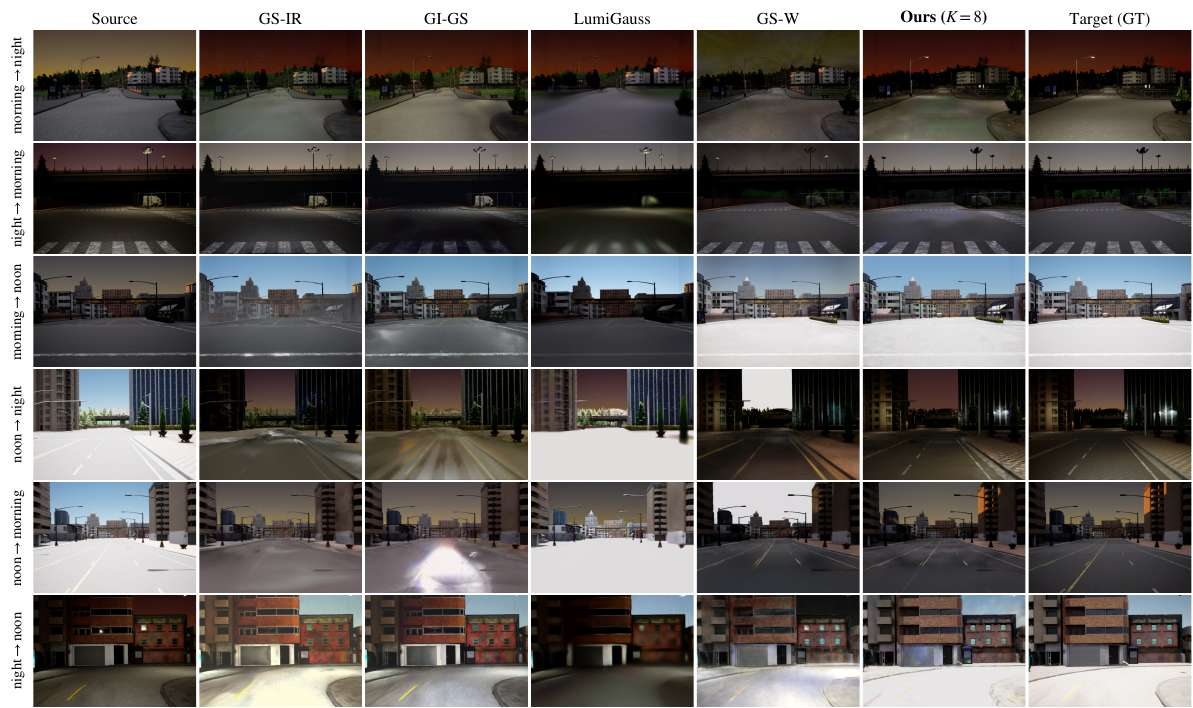}
\caption{Additional qualitative results for NVS on CARLA-ToD static scenes.}
\label{fig:appenndix_3}
\end{figure}

\begin{figure}[ht]
\centering
\includegraphics[width=\linewidth]{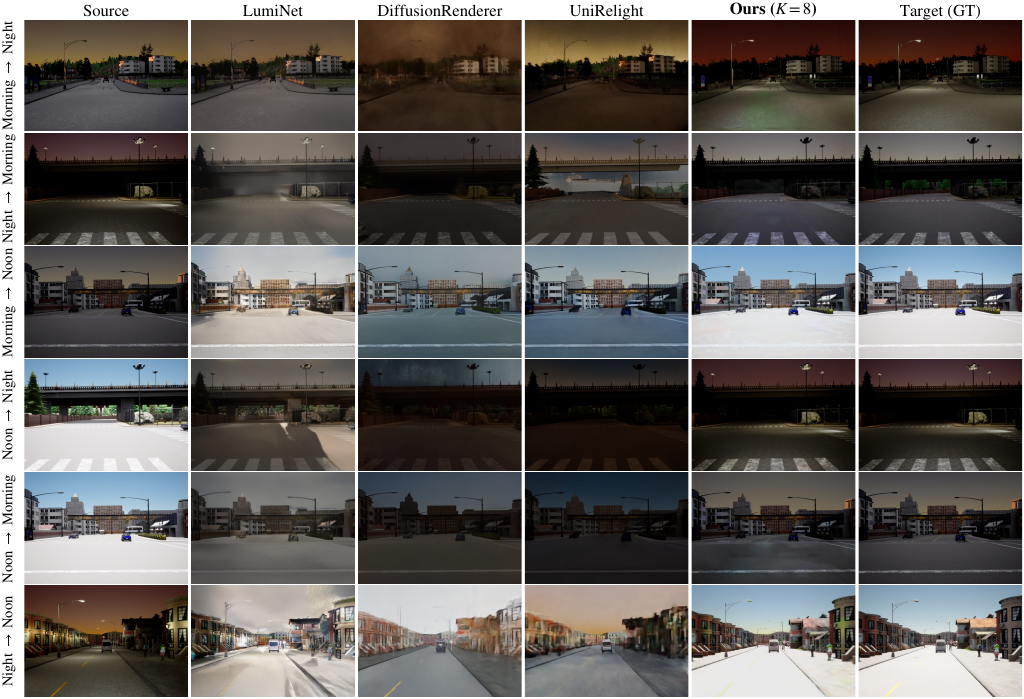}
\caption{Additional qualitative results for NVS on CARLA-ToD dynamic scenes.}
\label{fig:appenndix_4}
\end{figure}

\begin{figure}[ht]
\centering
\includegraphics[width=\linewidth]{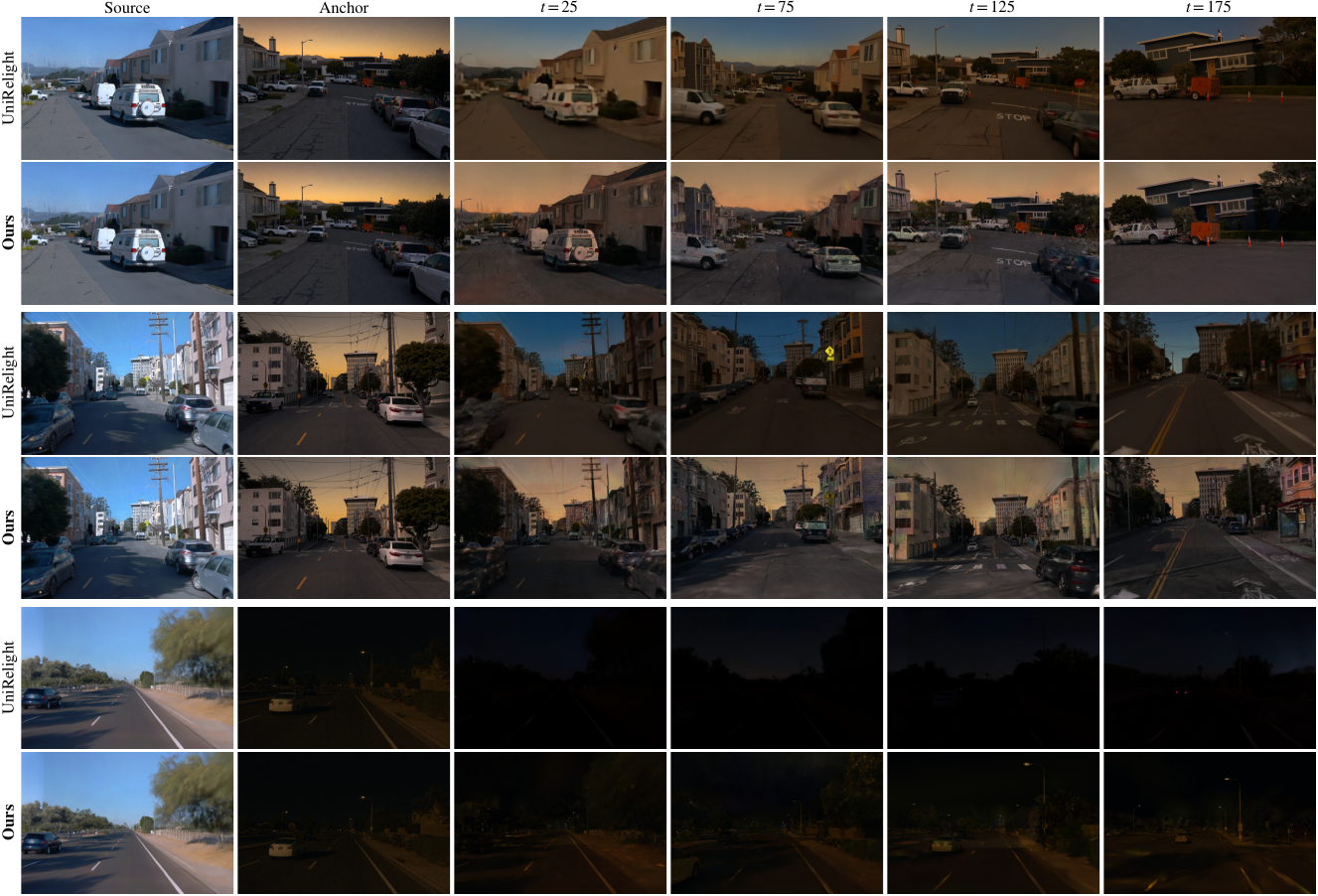}
\caption{Additional qualitative results for NVS on WOD.}
\label{fig:appenndix_5}
\end{figure}

~\Cref{fig:appenndix_3} and ~\Cref{fig:appenndix_4} present additional cross-sectional qualitative results regarding various segments of the CARLA-ToD Dataset.
~\Cref{fig:appenndix_5} shows the qualitative results from the Waymo Open Datasets.

\end{document}